\documentclass{article}
\usepackage[preprint]{neurips_2026}
\usepackage[utf8]{inputenc}
\usepackage[T1]{fontenc}
\usepackage{multirow}
\usepackage{comment}
\usepackage{hyperref,url,booktabs,amsfonts,amsmath,nicefrac,microtype,xcolor,graphicx}
\hypersetup{hidelinks}
\graphicspath{{figures/}}
\newcommand{\canonicalclips}{448}

\newcommand{\sweeptablerows}{%
Bounce & 5 & 6/8 & $-0.58$ ($p{=}0.31$) & 1/24 \\
Incline & 6 & 8/8 &  N/A & slide 0/24; stay 24/24 \\
Push & 5 & 7/8 & $-0.06$ ($p{=}0.83$) & 2/40 \\
Pendulum & 4 & 2/8 & $0.80$ ($p{=}0.20$) & 0/32 \\}

\newcommand{\nsweeptext}{22/22}     
\newcommand{\nsweepimgtext}{19/22}  

\title{One Model, Two Physical Stories: Auditing Misalignment in Multi-Modal World Modeling}
\author{%
  Geigh Zollicoffer$^{1}$ \quad
  Minh Vu$^{2}$ \quad
  Rajiv Ranasinghe$^{3}$ \quad
  Manish Bhattarai$^{1}$ \\[0.5em]
  $^{1}$Theoretical Division, Los Alamos National Laboratory \\
  $^{2}$Computing and Artificial Intelligence Division, Los Alamos National Laboratory \\
  $^{3}$Earth and Environmental Sciences Division, Los Alamos National Laboratory \\
  Los Alamos, NM 87545
}

\begin{document}
\maketitle

\begin{abstract}
World models, systems that generate what happens next given current environmental conditions, are increasingly being implemented with multi-modal generation in mind. 
However, generating multiple modalities simultaneously, such as visual simulations alongside physical state predictions in the form of text, introduces the risk of cross-modal inconsistency.
Tested separately, both outputs may look
convincing while still disagreeing: a model can calculate that a ball
should rebound in one modality, then generate no rebound in another modality, to say nothing of diverging from real-world dynamics entirely. In this work we focus on two failures explicitly: \emph{Internal
misalignment}, the disagreement between the world model's generated video and the same
world model's prediction in a different modalities, and \emph{external misalignment} the disagreement
between the world model's generation and an analytic physical environment. We derive common contracts of event, magnitude, timing, and construct a physics grounded pipeline to make comparisons measurable in both external and internal settings. We then ask
whether progressively supplying the model's own contract (the A ladder for the internal setting) or a
corrected physical contract (the B ladder for the external setting) closes the respective gaps. Across
four mechanisms and 20 settings, we find that while language answers all 22 text probes correctly with respect to the true environment,
the neutral video is often in disagreement, suggesting that the current unified backbones may not be capable of correct reasoning, internal consistency, and external physical fidelity all at once.
\end{abstract}

\section{Introduction}
Imagine asking a model how high a dropped ball should rebound, and it answers
``2.7 diameters.'' Give the same scene and parameters to a different modality of a system, and the ball may not rebound at all. An issue arises when interfaces belong to one
shared backbone; both outputs may look individually convincing; together they tell
incompatible physical stories.

This inconsistency is consequential. A world model used for robot planning,
counterfactual evaluation, or synthetic data generation must preserve the
effect of an intervention, not merely produce a realistic-looking future. A
planner can exploit a video model that ignores friction, delay, or contact
magnitude even when a language interface can state the governing relation.
Conversely, a correct verbal answer is not evidence that the sampled future
obeys it.

\begin{figure*}[t]
\centering
\includegraphics[width=0.98\textwidth]{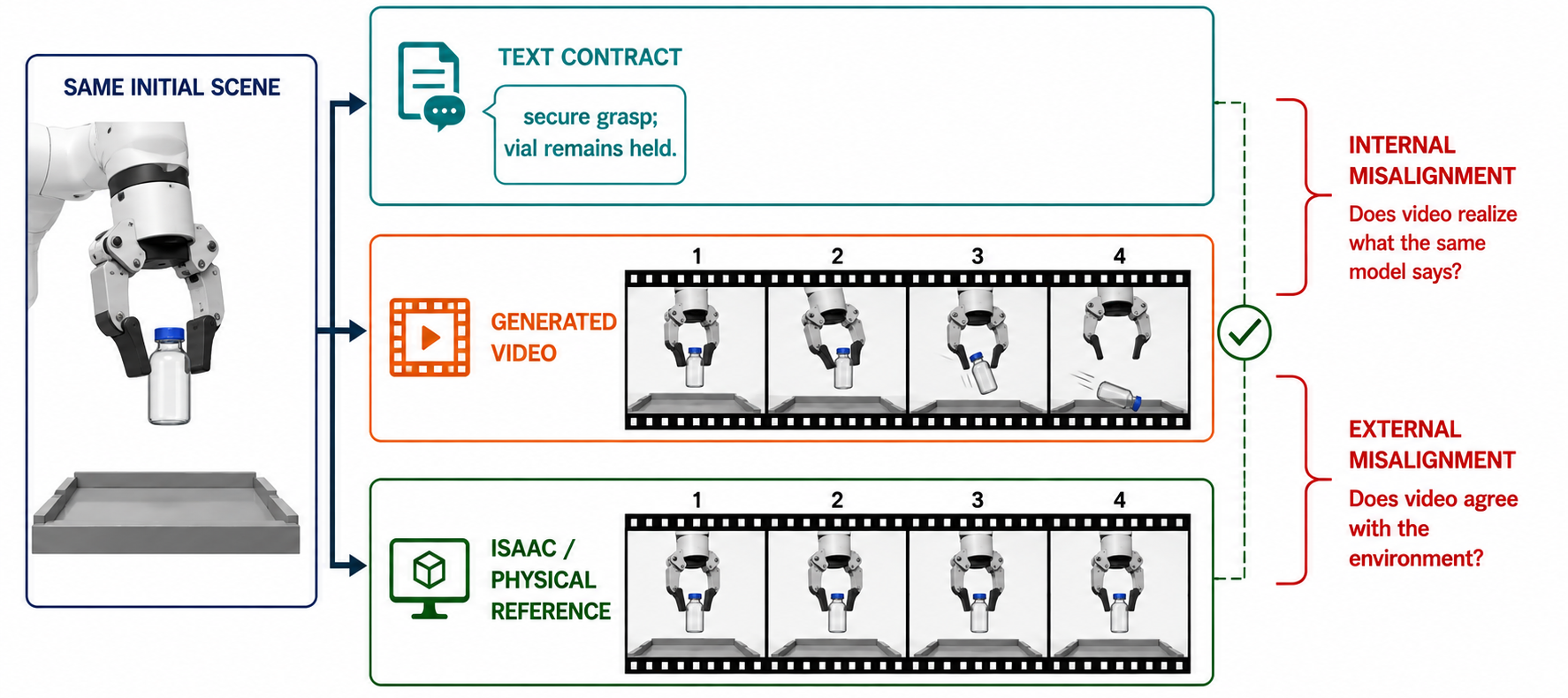}
\caption{\textbf{A schematic of one generated future, two alignment gaps.} From the same robot scene, text predicts a secure
grasp and the Isaac physical reference lifts the vial, while generated video
fails to realize the hold/lift. The video-text disagreement is internal misalignment; the
video-environment disagreement is external misalignment. Both the generated video and text contract are outputted from the same world model backbone.}
\label{fig:motivation}
\end{figure*}

The motivating robot case is grounded in an originating paired hold/slip check. The vial
tracker detected the object in every frame. Normalized vertical rise was
$0.448$ frame-heights for the Isaac hold (roughly centered in the video), but $0.004$ for the Cosmos hold (towards the bottom of the screen); the
Cosmos slip was similarly near-zero ($0.006$). 

We distinguish two alignment questions. \emph{Internal alignment} asks whether
the world model generated video realizes the same predictions as what the same backbone states through its language or vision-language modalities.
\emph{External alignment} asks whether video agrees with the analytic environment. These references answer different questions even
when they happen to predict the same outcome: one tests cross-modal
consistency, while the other tests physical correctness with the ground truth environment.

\begin{figure*}[t]
\centering
\includegraphics[width=0.98\textwidth]{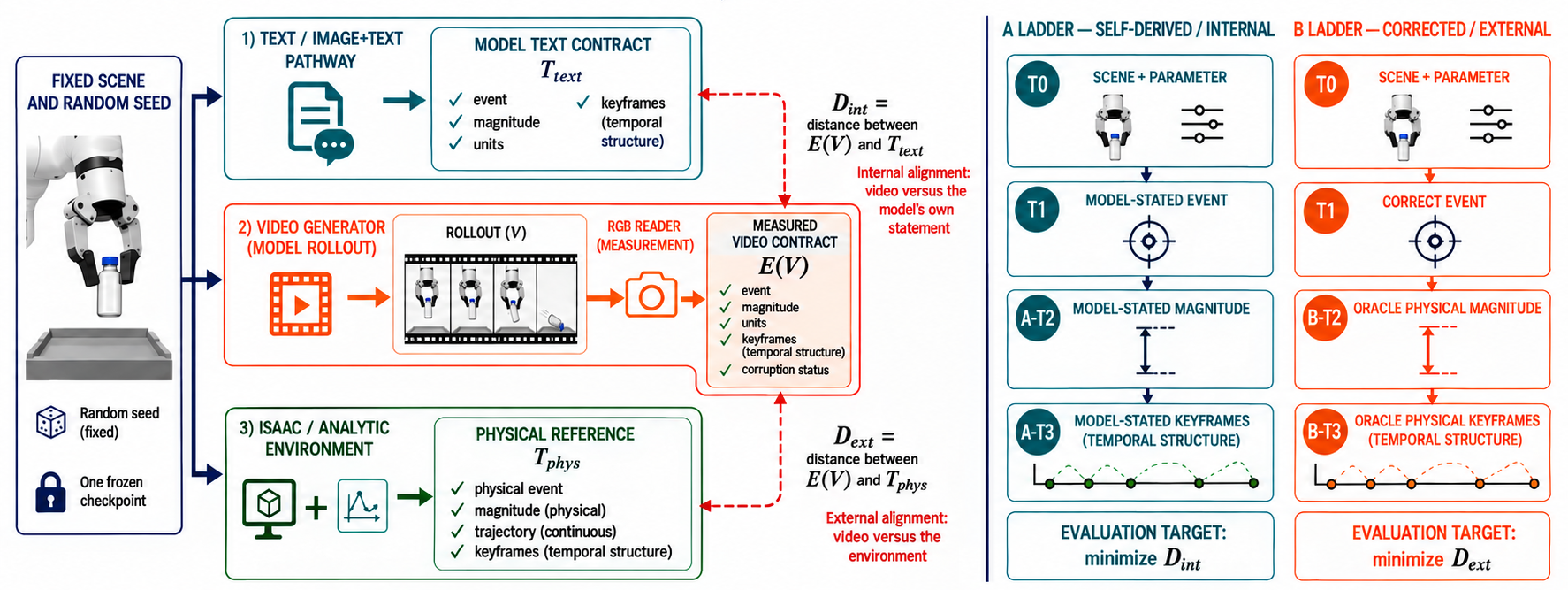}
\caption{\textbf{Two references and two tiered interventions (schematic).}
The same measured video contract $E(V)$ is compared with the model's text
contract for internal alignment and with Isaac/analytic physics for external
alignment. The A ladder adds self-derived event, magnitude, and keyframes; the
B ladder adds corrected physical information. 
}
\label{fig:hypotheses}
\end{figure*}

Our experimental flow follows that distinction. We first measure both gaps and then apply a tiered intervention, which we summarize in the form of two different \textit{ladders}. The \emph{A ladder} progressively
supplies the model's own text-derived magnitude and trajectory, testing whether
more internal information closes the video--text gap. The \emph{B ladder}
supplies corrected physical magnitude and trajectory, testing whether more
oracle information closes the video--environment gap (see Figure~\ref{fig:hypotheses} for a full visualization).

Concretely, our work aims to investigate four main questions about the world models generated video: 
\begin{itemize}
    \item \textbf{Q1 (External alignment):} Does the generated video match
$T_{\mathrm{phys}}$, the environment reference? 
\item \textbf{Q2 (Internal
alignment):} Does the generated video match $T_{\mathrm{text}}$, the same model's explicit
prediction? 
\item \textbf{Q3 (A ladder):} Does progressively adding
$T_{\mathrm{text}}$ reduce internal error of the generated video? 
\item \textbf{Q4 (B ladder):} Does
progressively adding corrected $T_{\mathrm{phys}}$ reduce external error of the generated video?
\end{itemize}


To measure alignment against the world model's generated video, we measure and define each modality's explicit, verifiable prediction i.e its \emph{predictive
contract}. From the text, we extract a binary event (e.g., bounce/no bounce or slide/no slide), an event magnitude (e.g., bounce height), and, where appropriate, keyframes and timestamps specifying when and where an object should appear. From the video, we recover the object’s pixel trajectory and express it in the same object-relative units used by the text-derived predictions. Ground-truth measurements are provided by the simulation environment. We use Isaac/PhysX because it provides a controllable external environment in which we can systematically vary object properties, spatial configurations, and physical parameters, such as mass, distance, and static friction.


\textbf{Contributions.}
(1)~We formalize internal video–text alignment and external
video–environment alignment with predictive contracts
(\S\ref{sec:framework}). (2)~We construct a four-mechanism intervention testbed
and validated RGB reader to measure both gaps. (3)~We introduce an A/B tiered
testing design that evaluates whether progressively supplying self-derived or
oracle physical information can close the corresponding gap. (4)~We
substantiate the resulting claims through canonical scoring of
\canonicalclips{} clips, sensor-validity checks, and multi-seed evaluation.

\section{Related Work}
\textbf{Physical evaluation of video generators.} Benchmarks score generated
video for physical plausibility and commonsense
\citep{bansal2025videophy,li2025worldmodelbench}, or against real recorded
physical experiments with explicit auditing of prompt and ground-truth
confounds \citep{motamed2026physicsiq}. These establish \emph{that} generators
violate physics; they compare a generator to an external reference. We add
two things such benchmarks cannot express: a within-checkpoint comparison of
the generator against the \emph{same system's} explicit reasoning, and
one-parameter interventions that test directional response rather than
pointwise plausibility. Our result that most seeds retain one outcome class
across three parameter sweeps is about the sampled conditional distribution,
not any single clip's realism or universal model behavior.

\textbf{Understanding--generation gaps.} GapEval directly quantifies
bidirectional understanding--generation consistency in unified multimodal
models \citep{wang2026gapeval}. Our study is narrower in model breadth and
broader in physical instrumentation: it holds the scene fixed, intervenes on a
known parameter, and compares pixel trajectories with both an external
reference and the model's explicit prediction. The present A/B internal-target
experiment is diagnostic rather than an equivalence result (Q3 and Appendix).

\textbf{Multi-modal world models.} Cosmos 3 couples an autoregressive
understanding stream with a diffusion generation stream in one jointly-trained
checkpoint \citep{nvidia2026cosmos3}; V-JEPA~2 and related work frame video
models as predictors for understanding and planning \citep{assran2025vjepa2,
mazzaglia2024genrl}. Our architectural audit (\S\ref{sec:testbed}) uses the
released modeling code to establish exactly which parameters the two pathways
share, which is what licenses the ``reasoning is available to the generator,
yet not realized'' framing.

\textbf{Steering and aligning generated physics.} Force prompting, physics
priors, inference-time alignment with latent world models~\citep{hafner2024masteringdiversedomainsworld, zollicoffer2025novelty}, and concept-vector
steering all improve or control physical behavior in generated video
\citep{gillman2025force,narayanan2026phyco,yuan2026physicsalignment,
alam2026causalphysics,joseph2026interpreting}. Our crossed correct/incorrect
directive design is complementary: before steering, it measures what the
text channel actually controls, event occurrence, in both directions,
and what it does not, magnitude on an unmodified checkpoint.

\section{Physics Fidelity Framework}\label{sec:framework}
\begin{figure*}[t]
\centering
\includegraphics[width=0.98\textwidth]{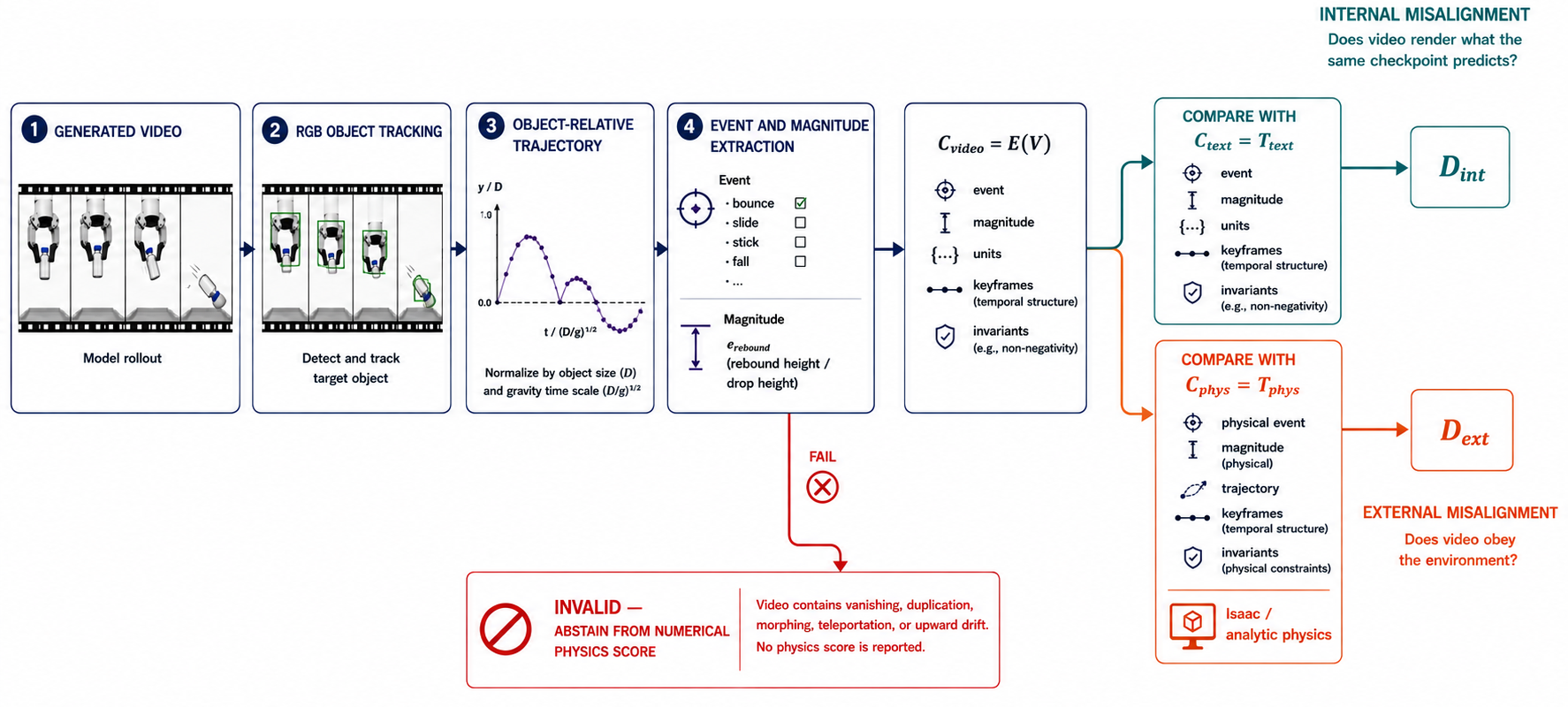}
\caption{\textbf{One measured video contract, two alignment tests
(schematic).} The frozen RGB reader maps a rollout to event, magnitude, units,
keyframes, and invariants. Comparing that same contract with $T_{\mathrm{text}}$
yields $D_{\mathrm{int}}$; comparing it with $T_{\mathrm{phys}}$ yields
$D_{\mathrm{ext}}$. Vanish, duplicate, morph, teleport, and non-physical-drift
failures are rejected rather than assigned a numerical score.}
\label{fig:measurement}
\end{figure*}

Let \(x\) denote an initial scenario comprising of a conditioning frame and a scene description with grounded physical measurements, including, but not limited to, object mass or weight, object dimensions, distances between objects or surfaces. Let \(\theta\) denote a physical parameter, such as the coefficient of restitution \(e\), coefficient of friction \(\mu\), launch speed \(v\), or pendulum length \(L\); and let \(V(x,\theta,p)\) denote a video sampled from the generation pathway using prompt \(p\). A reference simulator or governing physical equations provide $T_{\mathrm{phys}}(x,\theta)$, the physically expected trajectory of the tracked object under scenario $x$ and physical parameters $\theta$. Before evaluation, we verify that this trajectory can be determined by the specified initial conditions and governing dynamics defined by scenario $x$ (see Appendix~\ref{app:example_verification} for an explicit inclined-plane example). In contrast, $T_{\mathrm{text}}(x,\theta)$ denotes the trajectory predicted by the model’s text pathways. A
calibrated pixel extractor $E$ maps videos to observable trajectories with distance converted to
\emph{object-relative units} (the object's own diameter), making measurements
camera-calibration-free (for an illustration see Figure~\ref{fig:measurement}).

Formally, we represent each trajectory using a common predictive contract
$C=(z,m,u,K,I)$,
where $z$ denotes the event class (e.g., slide/stay or bounce/no-bounce), $m$ the measurable continuous quantity of interest (e.g., displacement, amplitude, or period), $u$ its units, $K$ any observable keyframes or timestamps, and $I$ scene invariants such as object size and camera pose. The physical reference $T_{\mathrm{phys}}(x,\theta)$ and the model's text prediction $T_{\mathrm{text}}(x,\theta)$ are expressed under this same schema ($C_{text},C_{phys}$), while the video extractor $E_v$ maps the generated video $V$ to the corresponding observed contract $C_{\mathrm{video}}=E(V)$. Using a shared schema ensures that like quantities are compared across modalities; for example, displacement is compared with displacement rather than period, and that quantities that are missing, corrupted, or below the extractor's measurement resolution remain unscored.

\textbf{Reasoning fidelity} first checks whether $T_{\mathrm{text}}$ agrees
with $T_{\mathrm{phys}}$. Both alignment tests then use the same measured video
contract,
$C_{\mathrm{video}}=E\!\left(V(x,\theta,p)\right)$, while changing only the
reference:
\[
D_{\mathrm{int}}(x,\theta,p)
=d\!\left(C_{\mathrm{video}},T_{\mathrm{text}}(x,\theta)\right),\qquad
D_{\mathrm{ext}}(x,\theta,p)
=d\!\left(C_{\mathrm{video}},T_{\mathrm{phys}}(x,\theta)\right).
\]
Thus \textbf{internal misalignment} is disagreement with the same checkpoint's
text or image--text prediction, while \textbf{external misalignment} is
disagreement with the environment,
with mechanism-appropriate distances $d$, each on the observable the
extractor can read reliably from pixels: bounce, absolute error in
rebound apex (ball-diameters); incline, regime agreement (slide/stay) plus
displacement up to the visible-ramp ceiling (in cube-widths; in practice a binary
slide/stay result); push, pushed/not pushed; pendulum, absolute
error in swing amplitude (bob-diameters). A clip is tolerance-correct when
measurable and within $\max(0.5\text{ units}, 25\%)$ of the reference, or in
the correct regime for incline; these rules were fixed before scoring.
$D_{\mathrm{int}}$ asks whether the generator realizes \emph{what the model
itself predicted}; $D_{\mathrm{ext}}$ asks whether that realization is
physically correct. The $(D_{\mathrm{int}}, D_{\mathrm{ext}})$ plane separates
faithful-and-correct, faithful-but-wrong, unfaithful-but-correct, and unfaithful-and-incorrect.
When $T_{\mathrm{text}}=T_{\mathrm{phys}}$, the two distances coincide
numerically even though their interpretations remain different; separating
them empirically requires cases in which the two targets differ.

We then construct a tiered intervention (see Figure~\ref{fig:hypotheses} for an illustration) that incrementally adds context to the scenario $x$. Tier 0 supplies the
scene and parameter; Tier 1 adds the event. At Tier 2 and Tier 3:
the A ladder adds the model's own magnitude and keyframes and is evaluated by
$D_{\mathrm{int}}$, whereas the B ladder adds corrected physical magnitude and
keyframes and is evaluated by $D_{\mathrm{ext}}$. This paired construction asks
whether additional information closes the intended gap.

\textbf{Degeneration.} Generated videos can fail physically \emph{and}
ontologically: objects vanish, spawn, morph, or teleport, and frame geometry can
corrupt through letterboxing or other discontinuities, a class our tracker does
not gate and which we report by inspection. Every clip is gated
into \{\textsc{dynamic}, \textsc{static}, \textsc{degenerate}\}; only
measurable clips receive a number, and we report $P(\text{measurable})$,
$P(\text{correct}\mid\text{measurable})$, and an unconditional score treating
degeneration as failure.


\section{Experiments}~\label{sec:testbed}

\paragraph{World Model} Cosmos3-Super is one jointly-trained
Mixture-of-Transformers checkpoint with separate 31.2B understanding and
generation streams, 1.6B shared parameters, and two-way joint attention. Text,
image--text, and video modes load the same shards but execute largely distinct
stream parameters. During generation, prompt tokens pass through the
understanding stream and generation queries attend to its keys and values. Appendix~A.1 gives the full architectural and inference audit.

\paragraph{Tested Scenarios (see Appendix~\ref{app:visuals} for visualizations)}
\begin{description}\itemsep2pt
\item[\textnormal{\textbf{Bounce (restitution).}}] A ball is dropped onto a fixed
pedestal from height $h$ and rebounds to $e^{2}h$ for coefficient of restitution
$e$. Observable: rebound-apex height above the landing level, in ball-diameters.

\item[\textnormal{\textbf{Incline (Coulomb friction).}}] A cube rests on a
$25^\circ$ ramp and slides iff $\mu<\tan\theta$, accelerating as
$a=g(\sin\theta-\mu\cos\theta)$. Observable: distance travelled down-slope from the
start, in cube-widths.

\item[\textnormal{\textbf{Push (kinetic friction).}}] A cube on a flat table is
given an initial horizontal speed $v_0$ and, opposed only by kinetic friction,
decelerates at $a=\mu_k g$ and coasts to rest after $d=v_0^{2}/(2\mu_k g)$.
Observable: horizontal distance travelled before stopping, in cube-widths.

\item[\textnormal{\textbf{Pendulum (timing).}}] A bob on a rigid rod of length $L$
is released from $\theta_0$ and swings with period $T=2\pi\sqrt{L/g}$. Observable:
horizontal swing amplitude about the lowest point, in bob-diameters (with the
period read from zero-crossings).

\item[\textnormal{\textbf{Grasp (contact friction).}}] A Franka gripper closes on a
vial, which is lifted iff Coulomb friction overcomes gravity, $\mu\cdot\text{grip}
\ge mg$. Observable: vertical rise of the vial. An additional setting we also
consider.
\end{description}
\paragraph{Reference simulator} Isaac Sim 6.0.1 (PhysX) supplies controllable bounce,
incline, projectile, and pendulum scenes with one intervened parameter and one
pixel-measurable observable each (Appendix Table~\ref{tab:mechanisms}). 

\paragraph{Evaluator and validation} A conditioning-frame--locked tracker maps
every clip to event, object-relative magnitude, object-relative displacement, and corruption status
(Fig.~\ref{fig:measurement}). On six lossless Isaac feeds it recovers event
class on 6/6, and in our experiments RGB and sensor traces agree framewise
($R^2{=}0.989/0.997$). Appendix~A.3 reports sensor,
camera, metric-amendment, and adjudication details.


\paragraph{Protocol} For each (mechanism, $\theta$), the identical scenario is
posed to the text and image+text pathways (temperature 0), and the video
pathway generates from the same conditioning frame under a \emph{neutral}
prompt (sizes and parameter stated; outcome never stated). The tiered design
shares T0 (scene/parameter) and T1 (event), then separates the model-derived A
and corrected-physics B references at T2 (magnitude) and T3 (keyframes). The
paired final-script A/B arm contains 32 clips at four seeds. Because most
measurable $T_{\mathrm{text}}$ and $T_{\mathrm{phys}}$ targets coincide, it is
diagnostic rather than a powered comparison of ladder slopes.

\textbf{Wrong directives (W1, W2)} The wrong-directive arms replace the correct
language with a false statement of the outcome while holding the physical
parameter, conditioning frame, and duration fixed. W1 is
\emph{qualitative-wrong}, it asserts the event does not happen (``the ball does
not bounce,'' ``the bob remains hanging and does not swing''), and W2 is
\emph{magnitude-wrong}, commanding a false extent (e.g.\ a $0.3$-diameter rebound
where physics dictates $2.7$). Their purpose is to test the \emph{direction} of
language control: if the tier effect were the model recalling physics from the
parameter, a false description should not matter. 

\section{Results}\label{sec:results}
\subsection{Prerequisite: Is the text reference well defined and correct?}
Across
the 20-setting sweep plus two pendulum wording permutations, the text pathway is
exact on \nsweeptext{} probes (e.g.\ 0.0333 diameters at $e{=}0.1$, 31.9 diameters at
$v{=}5$), and image+text is \nsweepimgtext{} strict: all 19 parsed final
answers are correct over binary event prediction (verified with the environment directly; three responses exhaust a 4096-token budget).

\subsection{External and internal alignment under intervention}
Where text and physics declare the same observable and the text pathway is
correct, a neutral video that misses the reference simultaneously exhibits
external misalignment with $T_{\mathrm{phys}}$ and internal misalignment with
$T_{\mathrm{text}}$.

\begin{figure}[t]
\centering
\includegraphics[width=.7\linewidth]{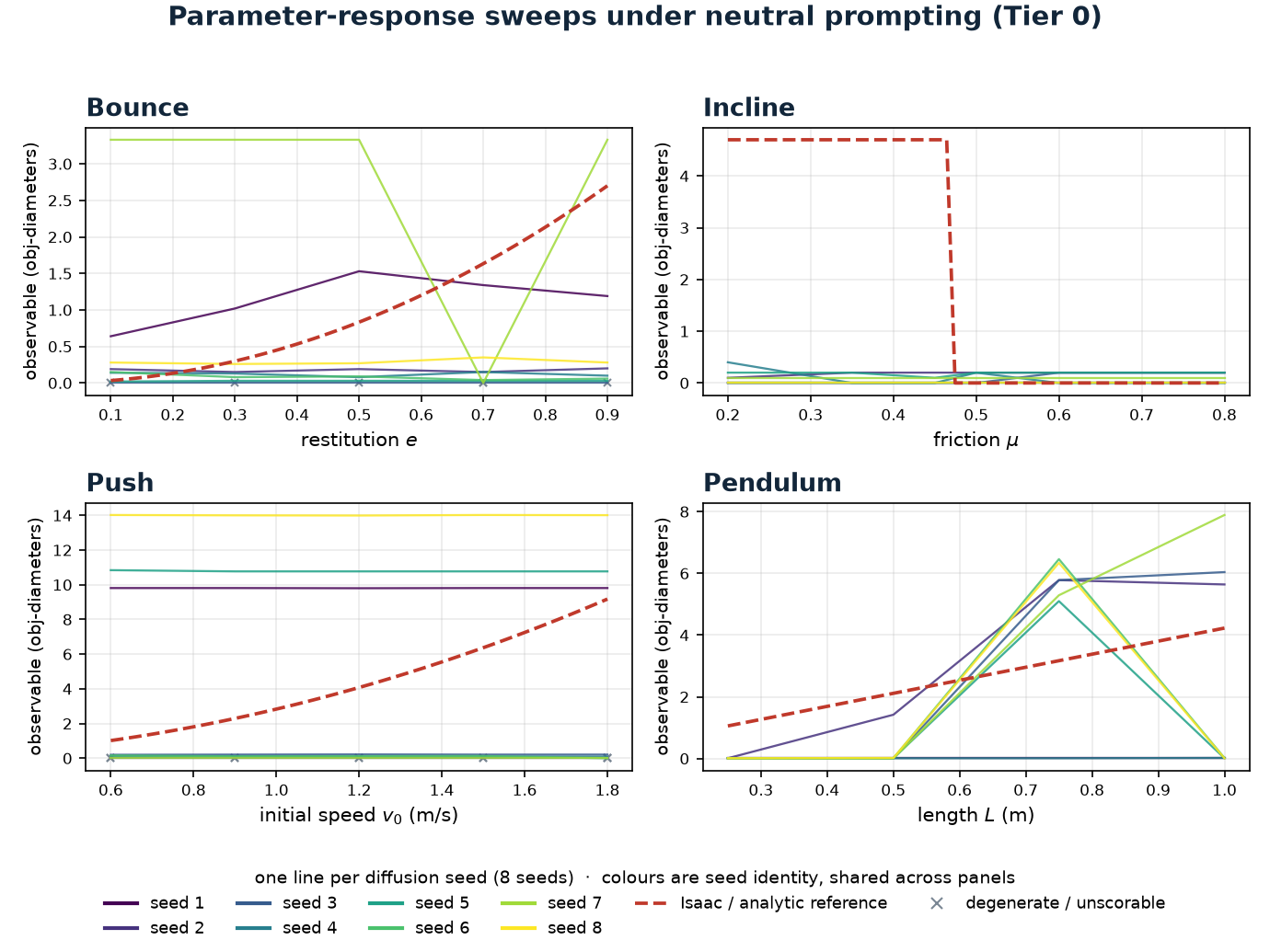}
\caption{\textbf{Parameter-response sweeps under neutral prompting (Tier 0).} Each
panel: one mechanism, $x$ = the main physical parameter stated in the prompt, $y$ =
the observable read from pixels in object-relative units; one line per
diffusion seed; dashed = Isaac reference (for incline, the visible-ramp
ceiling of the evaluator, 4.7 cube-widths, is drawn for the slide regime);
$\times$ = corrupted, degenerate, or unscorable clip. Eight seeds per setting.
Three mechanisms are flat and seed-striped; the pendulum swings
only at $L\geq0.5$, non-monotonically in $L$, with a 1.2--2$\times$
non-physical gain.}
\label{fig:sweep}
\end{figure}

Figure~\ref{fig:sweep} shows the parameter-response sweeps: 20 settings
(restitution $e\in\{0.1,\dots,0.9\}$; friction $\mu\in\{0.2,\dots,0.8\}$
bracketing the slide threshold $\tan 25^{\circ}{=}0.466$, at which PhysX
itself transitions between $\mu{=}0.45$ and $0.50$; launch speed
$v\in\{1.5,\dots,5.0\}$\,m/s; length $L\in\{0.25,\dots,1.0\}$\,m), eight seeds
each, 160 neutrally-prompted 81-frame clips, scored in one frozen pass with the
start box declared from each conditioning frame before scoring.

\textbf{Sensitivity to small prompt
edits} Changing the parameter value edits one token of a $\sim$120-token prompt at fixed
noise, and one might expect any such edit to leave the sample nearly unchanged. However, edits of comparable size
that name an \emph{event}, i.e ``the cube slides down the ramp'' vs.\ ``the cube stays in
place,'' ``swings back and forth'' vs.\ ``remains hanging'', flip the outcome class of
the same seeds (incline $0/8\to6/8$; pendulum $7/8\to3/8$), suggesting that insensitivity is not likely a cause of unresponsiveness. The conditioning channel is
responsive to words about what happens; it is unresponsive to the number that determines
what happens.

\textbf{Guidance affects swing occurrence but not amplification}
Regenerating the $L{=}1$ neutral condition on Pendulum with a change of guidance (a hyperparameter which controls sensitivity to the prompt during video generation) to 3.0 and 9.0 (four seeds
each) leaves seed 234 swinging with an over-large amplitude at all three
scales. Seed 345 swings at guidance 6.0 but freezes at 3.0 and 9.0; seeds 123
and 456 remain frozen. Thus the gain persists conditional on a rendered swing. In the over-amplified clips the bob's
pixel diameter varies by only 16--26\% across the swing (median 19--22\,px), while the
measured amplitude exceeds the physical maximum by $\sim$90\%; a depth excursion large
enough to inflate the amplitude that much would roughly double the apparent bob size.


\begin{table}[t]
\centering\scriptsize
\setlength{\tabcolsep}{3pt}
\caption{Parameter-sensitivity summary under neutral prompting (Tier 0, 8 seeds per setting).
Class-constant: seeds whose outcome class never changes across the sweep. Event-rate $\rho$:
Spearman of $P(\textsc{dynamic})$ vs.\ parameter across settings (undefined when constant).
Correct: unconditional tolerance-correct clips over settings above evaluator resolution.}
\label{tab:sens}
\begin{tabular}{lcccc}
\toprule
Mechanism & settings & class-constant seeds & event-rate $\rho$ & correct (uncond.) \\
\midrule
\sweeptablerows
\bottomrule
\end{tabular}
\end{table}

\subsection{Can corrected information close misalignment?}
We cross the physical parameter (Q1--Q2) with language that states a
\emph{correct} outcome---neutral T0 (parameter only), T1 (qualitative), T2
(magnitude), T3 (keyframed)---or a \emph{deliberately incorrect} one (W1
qualitative-wrong, W2 magnitude-wrong), holding parameter, conditioning frame,
and duration (81 frames) fixed across eight seeds; all $192$ clips ($4$ axes
$\times$ $6$ conditions $\times$ $8$ seeds) are scored in the canonical pass
(Table~\ref{tab:correctness}). We call a clip \emph{valid} when it is
\textsc{dynamic} and its observable lies within tolerance of the commanded (or
true) magnitude; \emph{corrupted or degenerate} otherwise.

On pendulum, W1
suppresses the swing in five of eight seeds ($3/8$ dynamic vs.\ $6$--$7/8$ under
every correct tier), and on incline W2's small commanded slide collapses
displacement to $1/8$. The wrong directives thus establish that the tier
unlocking is \emph{caused by the language content}, the generator follows the
stated outcome, in either direction, rather than by the physical parameter or
the conditioning image, which by themselves leave the event unrendered.

\textbf{Language controls whether the binary event occurs, in both directions.}
On incline at $\mu{=}0.2$ (truth: slide), neutral and T1 prompts leave the cube
frozen ($0/8$ each, CI $[0,0.32]$), T2 unlocks motion in $3/8$ seeds, and the
keyframed T3 prompt unlocks it in $6/8$ (CI $[0.41,0.93]$), saturating at the
evaluator's visible-ramp ceiling (4.7 cube-widths). On pendulum at $L{=}1$,
where six to seven of eight seeds swing under any correct-directive tier, the
wrong directive ``remains hanging, does not swing'' \emph{suppresses} the swing
in five of eight seeds (W1: $3/8$ dynamic, CI $[0.14,0.69]$, vs.\ $6$--$7/8$
under T0--T3). Neither effect is available to the physical parameter itself.

\textbf{Language does not control magnitude.} A commanded rebound height of $2.7$
diameters (T2) yields one valid bounce, four corrupted or degenerate clips, and
three low-amplitude events with overlapping observables (T2: $0.62$--$2.36$
diameters). Even an explicit ``does not bounce'' instruction (W1) produces two
valid bounces and four corrupted trajectories. A commanded swing of $4$
bob-diameters (T2) yields over-amplified swings ($7/8$ dynamic, median $7.0$).
The one suggestive magnitude effect is on incline, where a commanded
$15$-cube-width slide (T2: $3/8$ dynamic, up to $4.8$ cube-widths) elicits more
displacement than a commanded $2$-cube-width slide (W2: $1/8$, up to $1.1$); at
eight seeds this is a consistent direction with overlapping intervals, not a
calibrated response.

\textbf{Rendered motion is not rendered physics.} The push axis does move---$3$--$4$
of $8$ seeds render a slide at every tier---but the motion never obeys the commanded
magnitude: the median travel is a fixed ${\sim}10$ cube-widths (about $2.6\times$ the
true stopping distance) and is independent of the stated $v_0$, so only $0$--$1/8$
clips fall within tolerance. Even the magnitude-wrong directive W2 still slides
($4/8$), and $1$--$5$ of $8$ clips per tier are degenerate or unscorable. Bounce T1
and T3 (the most explicit correct directives) are corrupted or degenerate in $6/8$
and $7/8$ seeds.

\paragraph{Does a richer B ladder close the quantitative external gap?}
No tier is generally monotone in physical fidelity. Only incline shows a
dose--response ($0/8, 0/8, 3/8, 6/8$ slide), and this reaches the evaluator's
visible ceiling rather than the commanded magnitude. Bounce is non-monotone with
only one tolerance-correct clip; pendulum remains over-amplified; push renders
motion but stays parameter-blind, holding a fixed ${\sim}10$-cube-width slide
regardless of tier or commanded $v_0$ ($0$--$1/8$ within tolerance). Richer
specification therefore controls \emph{whether} some renderable motion occurs more
reliably than \emph{how much}, while sometimes increasing degeneration.

\begin{table}[t]
\centering
\small
\setlength{\tabcolsep}{4pt}
\caption{Event occurrence vs.\ correctness across eight seeds, scored two ways:
\emph{External} (against the true physics value) and \emph{Internal} (against the
model's own asserted value). \emph{Valid Event}: seeds (of $8$) that render any
valid event; \emph{Valid Magnitude}: seeds whose observable is additionally within
tolerance of the target ($\pm30\%$; incline counts reaching the visible-ramp
ceiling; push is stopping distance in cube-widths). Correct directives T0--T3 add
increasingly explicit correct language; wrong directives W1 (qualitative-wrong) and
W2 (magnitude-wrong) state a false outcome.}
\label{tab:correctness}
\begin{tabular}{ll cccccc c cccccc}
\toprule
& & \multicolumn{6}{c}{\textbf{External} (vs.\ physics)} & & \multicolumn{6}{c}{\textbf{Internal} (vs.\ model's claim)} \\
\cmidrule(lr){3-8}\cmidrule(lr){10-15}
Axis & & T0 & T1 & T2 & T3 & W1 & W2 & & T0 & T1 & T2 & T3 & W1 & W2 \\
\midrule
\multirow{2}{*}{Bounce}
 & Valid Event     & 2 & 1 & 4 & 1 & 4 & 4 & & 2 & 1 & 4 & 2 & 4 & 4 \\
 & Valid Magnitude & 0 & 0 & 1 & 1 & 1 & 2 & & 0 & 0 & 3 & 2 & 1 & 2 \\
\midrule
\multirow{2}{*}{Incline}
 & Valid Event     & 0 & 0 & 3 & 6 & 0 & 1 & & 0 & 0 & 3 & 6 & 0 & 1 \\
 & Valid Magnitude & 0 & 0 & 2 & 5 & 0 & 0 & & 0 & 0 & 2 & 5 & 0 & 0 \\
\midrule
\multirow{2}{*}{Push}
 & Valid Event     & 3 & 3 & 4 & 3 & 1 & 4 & & 3 & 3 & 4 & 3 & 1 & 4 \\
 & Valid Magnitude & 0 & 0 & 1 & 1 & 0 & 1 & & 0 & 0 & 1 & 1 & 0 & 1 \\
\midrule
\multirow{2}{*}{Pendulum}
 & Valid Event     & 6 & 6 & 7 & 7 & 3 & 7 & & 6 & 6 & 7 & 6 & 3 & 7 \\
 & Valid Magnitude & 3 & 2 & 1 & 1 & 1 & 1 & & 0 & 0 & 0 & 0 & 0 & 0 \\
\bottomrule
\end{tabular}
\end{table}



\section{Limitations}
Isaac/PhysX is simulation, not hardware truth. We test one model series, four
mechanisms, and eight sweep seeds. Appendix~A records the complete
qualifications.

\section{Conclusion}

We evaluate whether multimodal world models remain consistent across their own modalities and with external physical dynamics. Using shared contracts over event, magnitude, and timing, we compare generated video against both the model's text-derived prediction, \(T_{\mathrm{text}}\), and a physics-grounded reference, \(T_{\mathrm{phys}}\), separating internal from external misalignment. 

Across four physical mechanisms and 20 settings, the text pathway answers all 22 physics probes correctly with respect to the reference environment, yet neutral video generation frequently fails to realize the corresponding dynamics, showing that correct reasoning in one modality does not guarantee either cross-modal consistency or physical fidelity in another. Intervention ladders that progressively supply the model's own predicted contract or a corrected physical contract can alter behavior, but do not consistently close these gaps across mechanisms. 

These results suggest that current unified multimodal backbones may not be able to express a single coherent physical world model across modalities, and that evaluating world models requires testing not only whether each output is plausible in isolation, but whether their predictions agree with one another and with the environment they are intended to model.

\clearpage
\bibliographystyle{plainnat}
\bibliography{refs}

\appendix
\small
\section{Additional Analyses}

\subsection{Architecture and inference audit}
Cosmos3-Super has 64 layers at hidden size 5120. Each layer contains an
understanding stream (31.2B parameters; separate attention projections and
MLPs) and a generation stream (31.2B; separate MMDiT-style projections and
MLPs), with approximately 1.6B shared parameters, a 0.6B vision encoder, and a
0.7B video VAE. Text and image--text reasoning use temperature 0; video uses
UniPC at 30 steps, guidance 6.0, flow-shift 10, 24 fps, $640{\times}480$, and
49--81 frames. Released-code inspection verifies causal self-attention over
prompt tokens in the understanding stream and generation attention over
concatenated understanding/generation keys and values. This establishes a
route, not where quantitative content is lost. The video-encoder patch is a
numerically equivalent device-placement fix, verified on two input sizes.

\subsection{Scene and solver specifications}
The scenes are a 6\,cm, 0.2\,kg ball dropped 20\,cm onto a
$50{\times}50{\times}10$\,cm pedestal (restitution combine mode \emph{max});
a 6\,cm, 0.2\,kg cube on a 90\,cm ramp at $25^{\circ}$ with static and dynamic
friction equal (combine mode \emph{min}); a 6\,cm, 0.2\,kg cube given an initial
horizontal speed of 1.2\,m/s on a flat table with kinetic friction
$\mu_k{=}0.3$ (combine mode \emph{min}); and a 10\,cm pendulum bob released from
rest at $25^{\circ}$. Isaac Sim 6.0.1 uses a 60\,Hz physics step, the default TGS
solver, and no per-scene solver overrides. The originating Franka vial-grasp case
motivates the study but is not counted among these four controlled mechanisms.

\begin{table}[h]
\centering\scriptsize
\setlength{\tabcolsep}{3pt}
\caption{Controlled mechanisms, interventions, and pixel observables. Lengths
are measured in the tracked object's own diameter or width.}
\label{tab:mechanisms}
\begin{tabular}{llll}
\toprule
Axis & Scene & $\theta$ values & Observable \\
\midrule
Bounce & 6\,cm ball, 20\,cm drop & $e\in\{0.1,0.3,0.5,0.7,0.9\}$ & rebound apex ($e^2{\cdot}$drop) \\
Incline & 6\,cm cube, $25^{\circ}$ ramp & $\mu\in\{0.2,0.35,0.45,0.5,0.6,0.8\}$ & stay/slide; displacement \\
Push & 6\,cm cube, flat table ($\mu_k{=}0.3$) & $v_0\in\{0.6,0.9,1.2,1.5,1.8\}$\,m/s & stopping distance ($v_0^2\!/2\mu_k g$) \\
Pendulum & 10\,cm bob, $25^{\circ}$ release & $L\in\{0.25,0.5,0.75,1.0\}$\,m & amplitude (half peak-to-peak) \\
\bottomrule
\end{tabular}
\end{table}

\subsection{Test Setting Visualizations}~\label{app:visuals}
\begin{figure}[t]
    \centering
    \includegraphics[width=0.7\linewidth]{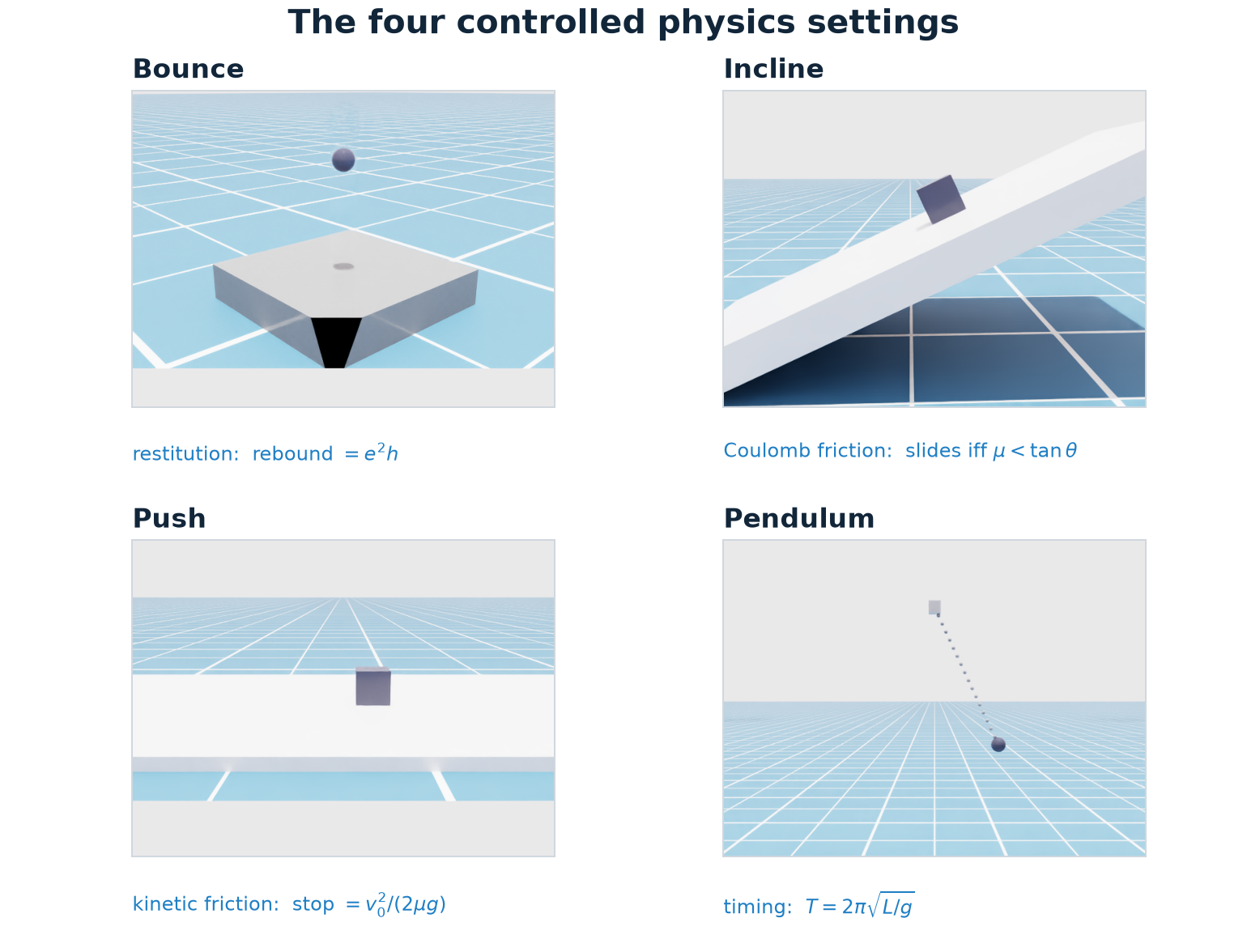}
    \caption{The four controlled physics settings, shown as their Isaac-twin
    conditioning frames. \textbf{Bounce} (restitution, rebound $=e^{2}h$):
    a ball released above a pedestal. \textbf{Incline} (Coulomb friction, slides
    iff $\mu<\tan\theta$): a cube on a $25^\circ$ ramp. \textbf{Push}
    (kinetic friction, stop $=v_0^{2}/(2\mu_k g)$): a cube given an initial
    velocity along a flat table. \textbf{Pendulum} (timing, $T=2\pi\sqrt{L/g}$):
    a bob on a rigid rod released from $25^\circ$. Each axis has one
    pixel-measurable observable in object-diameter units (rebound apex, down-slope
    displacement, stopping distance, swing amplitude).}
    \label{fig:settings}
\end{figure}

We evaluate four parameterized mechanics settings, each realized as an Isaac Sim
twin that supplies the exact ground-truth observable and a single conditioning
frame (Figure~\ref{fig:settings}). The four axes isolate distinct physical
mechanisms---restitution, Coulomb friction, ballistic motion, and oscillatory
timing---and each observable is expressed in the object's own diameter so it can
be read back from generated pixels and validated against exactly what the
conditioning specified.

\subsection{Explicit Validation Example}
\textbf{Example verification of \(T_{\mathrm{phys}}\): inclined plane.}
\label{app:example_verification}
\textit{Scenario:} Consider a cube of width \(w=0.06\,\mathrm{m}\) placed at rest at the top of a ramp of length \(0.90\,\mathrm{m}\) inclined at \(\alpha=25^\circ\), with coefficient of friction \(\mu=0.2\). \textit{Solution:} The cube slides when the downslope component of gravity exceeds the maximum frictional resistance, equivalently when

$$
\mu < \tan\alpha.
$$

Here,

$$
0.2 < \tan 25^\circ \approx 0.466,
$$

so the cube is expected to slide. Under the standard Coulomb-friction model, its downslope acceleration is

$$
a = g(\sin\alpha-\mu\cos\alpha)
   \approx 9.81(\sin25^\circ-0.2\cos25^\circ)
   \approx 2.37\,\mathrm{m/s^2}.
$$

Starting from rest, the expected displacement is therefore

$$
d(t)=\frac{1}{2}at^2.
$$

The cube reaches the end of the \(0.90\,\mathrm{m}\) ramp after

$$
t_{\mathrm{edge}}
=
\sqrt{\frac{2(0.90)}{2.37}}
\approx 0.87\,\mathrm{s}.
$$

Expressed in the object-relative units used by our video extractor, the total expected displacement is

$$
\frac{0.90}{0.06}=15
$$

cube-widths. Thus, for this scenario, \(T_{\mathrm{phys}}\) predicts a monotonic downslope trajectory reaching approximately \(15\) cube-widths of displacement at \(0.87\,\mathrm{s}\). We additionally verify this analytical prediction against the reference simulator's state trajectory, providing a direct check that the scenario specification and governing dynamics uniquely determine the physical reference used for evaluation.

\subsection{Original pilot and canonical replacement}
The originating four-point protocol crossed four mechanisms, six prompt cells
(T0, T1, A-T2, A-T3, B-T2, B-T3), and two seeds (123 and 42), for 48 clips.
Its validity gate found text--image+text agreement on all 7 comparable probes,
while video matched physics on only 1/8; repairing the ambiguous short-pendulum
wording yields the current 8/8 text and image+text prerequisite. In the
two-seed ladder, B-T3 produced a dynamic incline in 1/2 clips and a dynamic
pendulum in 2/2, whereas every post-T0 bounce clip was degenerate. Thus the
pilot suggested that explicit language could switch a coarse event without
establishing magnitude fidelity. The canonical study broadens the ladder to
eight seeds; its horizontal-translation axis is \emph{push} (slide-to-stop under
kinetic friction), which renders a slide but at a fixed, parameter-blind
magnitude---sharpening the same event-vs-magnitude dissociation on an axis that
does move.


\subsection{Evaluator qualification details}
The canonical reader uses per-axis color segmentation and a box declared from
the conditioning frame. Object loss yields \textsc{degenerate}; a rebound to
release height or sustained upward drift yields \textsc{corrupted}. The
incline camera exposes only 4.5 of 15 cube-widths, and the push camera's Isaac
feed is reliable only at its endpoints (replicator substeps corrupt the
intermediate frames), so its pixel scale is fixed from the release-to-rest
displacement (23.4\,px per cube-width, recovering the 3.91-cube-width truth).
We use pendulum half peak-to-peak amplitude because deviation from a temporal
mean is phase-biased; applying it to 110 clips changes 72 magnitudes but no event
class. Sensor checks find bounce apex error at most 13.8\%, correct incline
classification around $\tan25^\circ{=}0.466$, push stopping-distance error at most
4.0\% against the Isaac twin, and pendulum radius stability to
$2.1{\times}10^{-5}$\,m. Short-pendulum simulator amplitude is 58.5\% from
analytic truth (0.8\% at $L{=}1$), so analytic amplitude is the target.

\subsection{Stochastic realization}
With eight seeds per setting we can separate ``the model cannot produce the
correct dynamics'' from ``default sampling rarely selects them.'' Using the
tolerance rule of Q1--Q2 (degenerate and clamp-hit clips count as failures;
sub-resolution settings excluded), per-seed success@1 vs.\ an \emph{oracle}
best-of-8 that accepts a setting if any seed is correct: bounce $0.04 \to
0.33$, pendulum $0.00 \to 0.00$, incline $0.50 \to 0.50$ (the frozen output is
correct exactly in the stay regime, for every seed), push $0.05 \to 0.20$.
The oracle number is a diagnostic capacity bound: it uses simulator truth to
select and is not deployable. Thus a tolerance-correct bounce exists for one
third of scored settings, while no tolerance-correct pendulum appears in the
eight-seed support and push is tolerance-correct only at the highest speed
($v_0{=}1.8$\,m/s, where its fixed ${\sim}10$-cube-width slide happens to fall
within tolerance), and neutral prompting contains no incline slide.
Invalid-trajectory probability is $0.30$ (bounce), $0.28$ (push), and $0$
(incline, pendulum).

\subsection{Second checkpoint: Cosmos3-Nano}
We repeat the eight settings with the same frames, neutral prompts, 81-frame
length, canonical scorer, and four seeds (32 clips). Nano changes how often
motion appears, but not parameter sensitivity: bounce counts match across
$e{=}.3/.9$, incline slides at both $\mu{=}.2/.8$, push slides in $3/4$ seeds at
both $v_0{=}.6/1.8$ yet renders the same ${\sim}1.5$-cube-width displacement
regardless of commanded speed, and the long pendulum swings in $4/4$ seeds with
the same $1.2$--$1.8\times$ gain. This supports a shared failure family, not a
scale comparison. Three neutral paraphrases per dynamic setting (24 clips) alter
individual samples but not the seed-dominated conclusion.

\end{document}